# Using Semantic Web Services for AI-Based Research in Industry 4.0

Lukas Malburg[1] , Patrick Klein[1] , and Ralph Bergmann[1,2]

[1] Business Information Systems II, University of Trier, 54296 Trier, Germany
{malburgl,kleinp,bergmann}@uni-trier.de
http://www.wi2.uni-trier.de
[2] German Research Center for Artificial Intelligence (DFKI), Branch University of Trier, Behringstraße 21, 54296 Trier, Germany
ralph.bergmann@dfki.de

**Abstract** The transition to Industry 4.0 requires smart manufacturing systems that are easily configurable and provide a high level of flexibility during manufacturing in order to achieve mass customization or to support cloud manufacturing. To realize this, *Cyber-Physical Systems (CPSs)* combined with *Artificial Intelligence (AI)* methods find their way into manufacturing shop floors. For using AI methods in the context of Industry 4.0, semantic web services are indispensable to provide a reasonable abstraction of the underlying manufacturing capabilities. In this paper, we present semantic web services for AI-based research in Industry 4.0. Therefore, we developed more than 300 semantic web services for a physical simulation factory based on *Web Ontology Language for Web Services (OWL-S)* and *Web Service Modeling Ontology (WSMO)* and linked them to an already existing domain ontology for intelligent manufacturing control. Suitable for the requirements of CPS environments, our pre- and postconditions are verified in near real-time by invoking other semantic web services in contrast to complex reasoning within the knowledge base. Finally, we evaluate our implementation by executing a cyber-physical workflow composed of semantic web services using a workflow management system.

**Keywords:** Semantic Web Services · Industry 4.0 · Artificial Intelligence · Flexible Cyber-Physical Workflows · OWL-S · WSMO

## 1 Introduction

Nowadays, the industry is in a transformation towards the fourth industrial revolution, also known as Industry 4.0 in the German-speaking area [16]. The predominant application of *Artificial Intelligence (AI)* methods in *Cyber-Physical Systems (CPSs)* is a typical characteristic of this transformation [18]. In this context, flexibility is one of the central aspects for manufacturing companies particularly because of ever shorter market launch times and increasing customer demands for individualization [6,16]. In order to conduct close to reality Industry 4.0 research, we use a physical *Fischertechnik (FT)* simulation model because companies are often not willing to provide data from and access to their production lines for research purposes. To use AI applications in practise, knowledge must necessarily be available in formal and machine-readable representations [9]. *Semantic Web Services (SWSs)* address the problems of automatic discovering, composing, and executing by providing a declarative, ontological framework for describing them. Using AI methods (e. g., automated planning [22,23], multi-agent systems for decentralized manufacturing control [7,28], *Case-Based Reasoning (CBR)* [26,27]) to enhance flexibility in cyber-physical production workflows [2,34] inevitably require such semantic annotations. Several related work (e. g., [17,29,30]) exist that already address these issues by using SWSs. However, the currently available approaches that use SWSs in the context of Industry 4.0 focusing only on specific aspects and do not consider the entire context of manufacturing environments. Furthermore, the complex reasoning within the knowledge base makes real-time execution and monitoring of manufacturing processes difficult. In particular, as we expect these problems already in our simulation factory, it is probably not possible to use these approaches in a real production setting. For this reason, this paper presents an approach for modeling SWSs that avoids complex reasoning and is therefore suitable for near real-time AI-based applications. We identify requirements from three exemplary AI use cases that should be fulfilled by the developed SWSs. In the following, Sect. 2 describes the layout of our used physical *Fischertechnik (FT)* simulation factory and presents use cases in which the application of semantic information provided by semantic web services could be important. Furthermore, related work concerning the use of web services in manufacturing and a domain ontology for physical simulation factories are presented. The developed semantic web services themselves are described in detail in Sect. 3 and evaluated in Sect. 4. Finally, a conclusion is given and future work is discussed in Sect. 5.

## 2   Foundations and Related Work

### 2.1   Semantic Web Services for Flexible Production

The paradigm of *Service-Oriented Architectures (SOAs)* can be used to achieve manufacturing flexibility as it is needed for Industry 4.0 mass customization [16] and reconfigurability by decoupling functionality from the underlying implementation and location [21]. For implementing SOAs, web service technologies can be used [10] and, in particular, in the context of AI, these are semantically enriched, resulting in *Semantic Web Services (SWSs)*. Therefore, semantic technologies such as *Web Ontology Language for Web Services (OWL-S)* [24] for expressing the meaning of the web service interface can be used. In general, this enables automatic discovering, composing, and executing that are required for using AI methods.

Several works propose the use of SWSs for smart manufacturing in Industry 4.0 but focusing only on partial aspects and do not consider the entire context of the shop floor. For instance, Puttonen et al. [29] present an approach to use SWSs for executing manufacturing processes by means of three software agents represented as web services. One of these agents, referred to as Service Monitor, is a specialized web service that carries out semantic web service composition by using planning techniques w.r.t a given production goal and the current state of the world that is provided by a domain ontology. Therefore, they use an OWL ontology for describing the state of the production system as well as OWL-S and SPARQL expressions for semantically describing the available web services that offer production capabilities. In contrast, our work uses a physical instead of a virtual model and focuses on semantic descriptions of the underlying web services that are directly accessible in order to control manufacturing resources. Furthermore, we combine our SWSs with a comprehensive domain ontology of our production environment, whereas their modeled domain knowledge is limited to product definitions. Our work also focuses on a cyber-physical environment in which real-time sensor values are retrieved as part of service pre- and postconditions. Moreover, they just have two OWL-S processes, which result in 126 process variants after considering all possible permutations of descriptions.

Since modern production systems such as CPSs consist of many different components and therefore many stakeholders are involved in their development process up to the later use in the manufacturing of products, Lobov et al. [20] investigated the application of SWSs for orchestration of a flexible control. They propose OWL for modeling a Process Taxonomy, Product Ontology, Equipment Ontology, and Service Ontology and mainly discuss the responsibilities of involved persons for knowledge acquisition and maintenance rather than present their detailed semantic specification.

More recently and most similar to our work, Cheng et al. [6] presented an architecture and knowledge model for the integration of web services for flexible manufacturing systems. Their model includes ontologies that are similar to Lobov et al. [20]. However, the web services themselves are not semantically modeled, while Puttonen et al. [29] develop semantic web services but do not link them to a comprehensive domain ontology of the manufacturing environment. Based on our experiments, we assume that the continuous reasoning to evaluate the pre- and postconditions is too complex and time consuming for real world applications due to ongoing updates of the state of a CPS. Thus, there is a lack of research regarding the integration of semantically enriched web services with an existing domain ontology of a manufacturing environment to make production control more flexible as well as of research that considers the reasoning complexity in real-time applications sufficiently.

### 2.2 Industry 4.0 Simulation Factory Model

Similar to Cheng et al. [6], we use a physical *Fischertechnik (FT)* factory model for the simulation of an Industry 4.0 manufacturing environment. Such models are referred to as *Learning Factories* [1] and are used for education and Industry 4.0 research purposes (e. g., [4,13]). They are useful to investigate developed research artifacts under laboratory conditions to assess their suitability before they are potentially used in practice. Our factory consists of two identical shop floors that are linked for the exchange of workpieces as shown in Fig. 1. Each shop floor consists of four workstations with six identical machines: a sorting machine with color detection, a multi-processing workstation with an oven, a milling machine, and a workstation transport that connects both of them, a high-bay warehouse, and a vacuum gripper robot. Additionally, each shop floor has one individual machine, i. e., a punching machine in the first and a drilling machine in the second shop floor. Each shop floor is equipped with several light barriers, switches, and capacitive sensors for control purposes. Additionally, the first shop floor is enhanced with dedicated sensors such as acceleration, differential pressure, and absolute orientation sensors. RFID reader/writers are integrated in workstations on both shop floors and in the high-bay warehouse resulting in 28 communication points. This allows each workpiece to be tracked and the required manufacturing operations and parameters to be retrieved and adjusted during production as necessary. Furthermore, a camera is placed above the two shop floors to detect and track the workpieces.

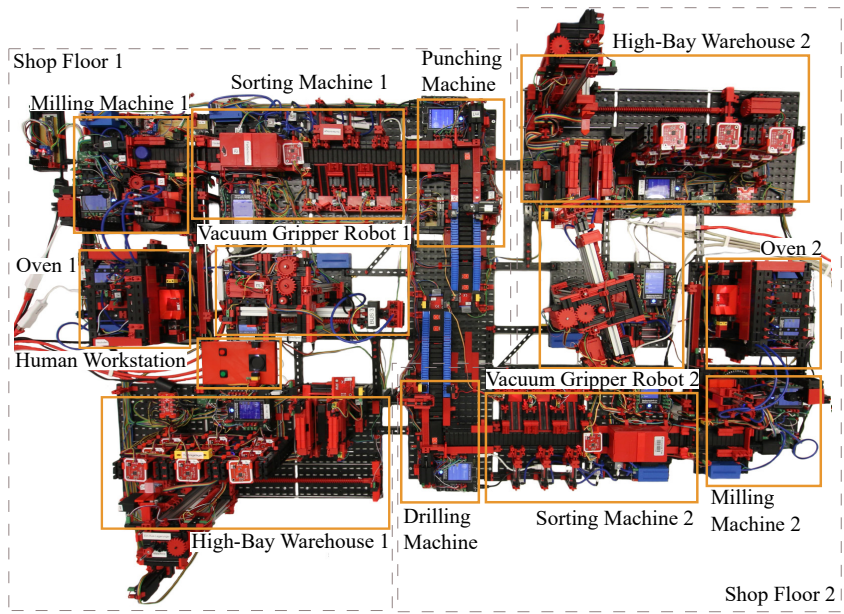

**Fig. 1.** The Two Fischertechnik Factory Simulation Models.

The availability of a knowledge representation in form of an ontology that represents the manufacturing environment enables engineers to share, reuse, and make knowledge in formal and machine-readable way explicit. For Fischertechnik simulation factories, FTOnto [14] provides a semantic description of resources such as sensors and actuators, products and raw materials, as well as operations such as manufacturing capabilities and handling. Each physical part of the factory is represented as an individual and is ordered in a sub-class hierarchy based on an established ontology for modeling a manufacturing system called MASON [19]. Additionally, object properties are used to model relations between individuals. Moreover, relationships between sensors and actuators are modeled by the SOSA ontology [12]. Figure 2 depicts a part of the domain ontology FTOnto. Classes are surrounded by an orange circle and instances by a purple rectangle.

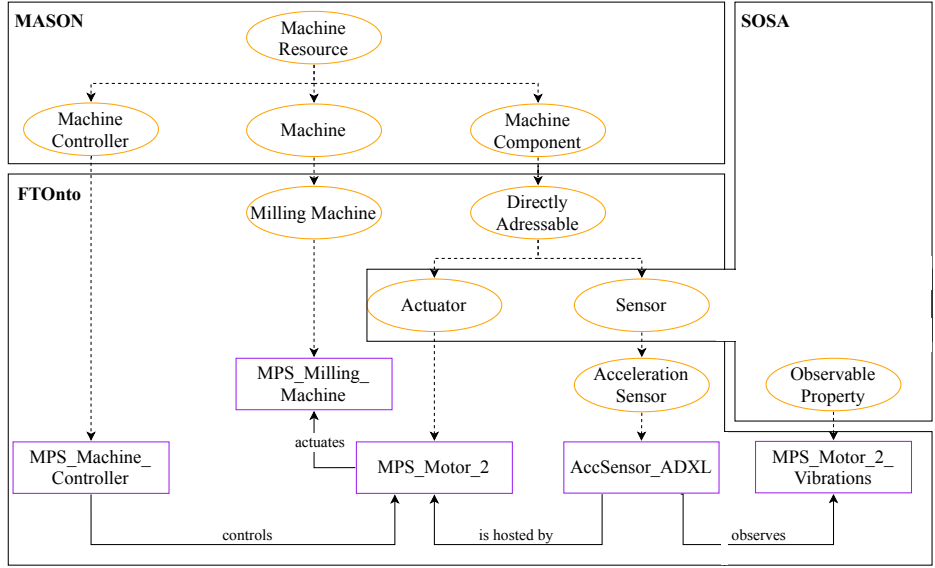

**Fig. 2.** Subset of Domain Ontology FTOnto.

The dashed arrow between both states a subclass relationship or that the instance is from the type of this class. For example, *MPS_MillingMachine* is the instance of the class *MillingMachine*. The statement that *MPS_MillingMachine* is driven by *MPS_Motor2* is modeled by the property *actuates*.

### 2.3 Use Cases

We now introduce three typical use cases by which we want to demonstrate the potential of semantic web services for AI-based research in the context of CPSs.

**Multi-Agent Systems for Decentralized Control** Agent-based approaches for intelligent manufacturing can be classified into functional and physical decomposition [35]. In the case of functional decomposition, an agent encapsulates functions, e. g., order processing, product design, or production planning. For instance, a planning agent either uses manually predefined plans that specify manufacturing operations and their sequence or generates a plan under consideration of the semantic descriptions of semantic web services that represent the manufacturing capabilities in order to achieve a given production goal such as a specific customer request [7]. The physical decomposition design approach results in agents that represent physical manufacturing resources or aggregations of resources and manufacturing operations. We consider a use case in which *Multi-Agent Systems (MASs)* utilize the web service infrastructure as an abstraction of physical devices to obtain information from the shop floor and to control the physical devices by invocation of web services.

**Workflow Planning of Manufacturing Processes** Similar to agent-based planning approaches, automated planning techniques can be used to build production workflows, i. e., a sequence of actions, from scratch or parts of them by using a complete domain model, an initial state, and a goal state [22,23]. Semantic web services enriched with preconditions and effects can be used for workflow planning. The work of Chen and Yang [5] is one of the first in this area that introduces the basic principle. Puttonen et al. [29] also use semantic web services and transfer the semantic descriptions into the *Planning Domain Definition Language (PDDL)* [25] for planning of manufacturing processes. Similar to the latter approach, we propose a use case in which automated planning techniques can be used to generate cyber-physical production workflows enabled by the definition of semantic web services and the transformation into PDDL.

**Business Process Management for Cyber-Physical Systems** The application of *Business Process Management (BPM)* techniques in cyber-physical environments can lead to numerous advantages. These are mainly due to the automatic retrieval of sensor data or events and are useful in determining the status of activities, making decisions about future process flow, and detecting deviations at an early stage [11]. For our use case, we consider that an intelligent system for BPM needs to access this data in different process steps or within a specified time frame in order to use it properly. For example, *Process-Oriented Case-Based Reasoning (POCBR)* [26,27] has shown great potential as an experience-based activity in similar research fields (e. g., in the cooking domain to represent cooking recipes as workflows [27] or in the domain of scientific workflows [37]) by using best-practise workflows from a case base. This approach can perhaps also be used to increase workflow flexibility in CPSs.

## 3   Semantic Web Services for AI-Based Research

In this chapter, our concept of semantically enriched web services for applying them in the use cases previously described is presented. The development process of our semantic web services follows the well-known ontology development methodology by Sure et al. [36] that contains the four steps *Kickoff*, *Refinement*, *Evaluation*, and *Application & Evolution* and that can also be applied for developing semantic web services. The steps are described in detail in the following.

### 3.1 Requirements

In this section, we introduce as part of the *Kickoff* phase *requirements (RQs)* that semantic web services should meet in order to be used for AI-based research activities in Industry 4.0. The requirements have been primarily derived from the use cases presented in Sect. 2.3. We do not divide the requirements according to these use cases, because some requirements are essential for more than one use case.

**RQ 1 – Provide Interoperability and Interconnectivity.** The services should be developed to achieve interconnectivity and interoperability from several controller types and programming languages [3,29].

**RQ 2 – Connect to Existing Knowledge Representations.** The services should be linked to an existing knowledge representation in form of a domain ontology. Domain ontologies are important for modeling production systems and for the transformation towards Industry 4.0 [29]. Due to the connection of semantic web services with an existing ontology, it is possible to directly check the result of a web service that returns sensor values for plausibility. Furthermore, it is possible to identify services that are currently not available due to an error of a physical manufacturing resource.

**RQ 3 – Enable Ontology and Knowledge Base Updates and Real-Time Verification.** In the context of Industry 4.0, it is necessary to use real-time data to make proper decisions [3,16]. Thus, it is necessary to keep the used ontology and its corresponding knowledge base up to date [20].

**RQ 4 – Abstract from Low-Level Control Commands.** It is important to abstract the web services from individual low-level control commands to create a hardware abstraction layer for CPSs. Therefore, each service should perform an atomic operation [31]. It is a necessary requirement to be able to break down end-to-end processes (see challenges 6 and 7 in [11]).

**RQ 5 – Enrich Web Services with Semantic Descriptions.** Due to the semantic enrichment of web services, it is easier to use AI-based technologies for improving resource utilization and for controlling the execution of the whole manufacturing process, i.e., preconditions and effects can be used to determine the impact of certain activities (see challenges 15 and 16 in [11] and [8]).

**RQ 6 – Model Relationships between Web Services.** Relationships between web services should be modeled to determine dependencies between services and to identify semantically similar services.

**RQ 7 – Parameterization of Web Services.** Reconfigurability of production systems is needed to satisfy changing process goals. By parameterizing web services that represent the possible configuration settings of physical manufacturing devices, different production goals can be achieved [20].

**RQ 8 – Resolve Mutual Exclusion.** To prevent multiple access to a single physical manufacturing resource, only one service may have simultaneous access to this resource to execute the corresponding operation [20]. During this time, however, further service requests should not be lost.

**RQ 9 – Orchestrate and Composite Web Services.** It should be possible
to orchestrate or composite the developed services to more complex processes
[20] and in order to fulfill an overall process goal [29,31].

### 3.2 Architecture Overview

Our concept integrates semantic web services with a domain ontology in an architecture of a CPS, which is, in fact, the foundation for research on AI-based control for flexible manufacturing processes. More precisely, we adopt the layered architecture for managing cyber-physical workflows proposed by Marrella et al. [23] depicted on the right side as a basis to define SWSs as a service layer. Additionally, we linked the SWSs of the service layer to an existing domain ontology of a manufacturing environment and to OWL-S. An overview of the proposed architecture is shown in Fig. 3. Starting from the bottom, the

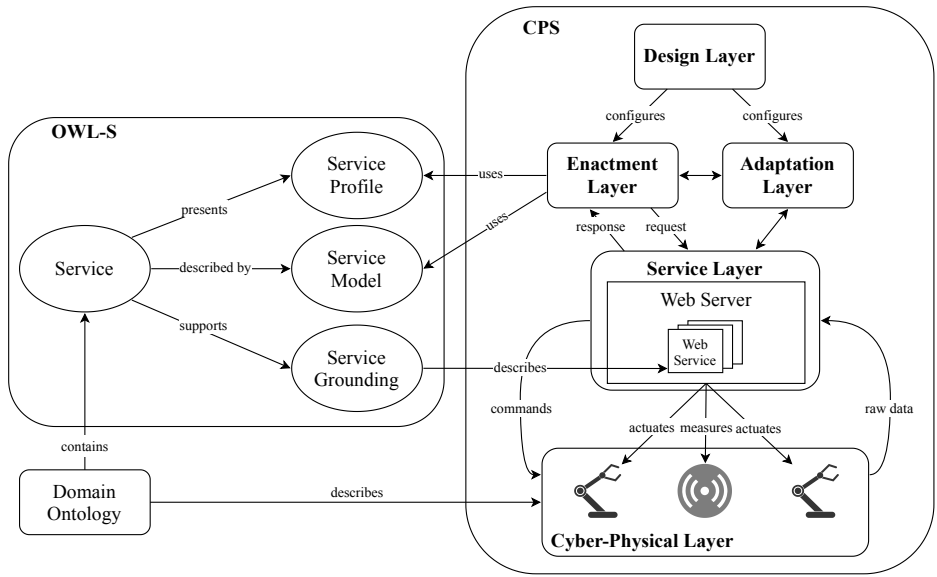

**Fig. 3.** Service Conceptualization for Flexible Production Control (Based on Marrella et al. [23]).

*cyber-physical layer* represents the manufacturing environment where several cyber-physical production workflows are executed. We propose to represent each resource (e. g., actuator, sensor etc.) as well as their relationships in a domain ontology. On top of this layer, the *service layer* is responsible for transmitting and executing control commands to the cyber-physical layer and receiving raw data from sensors. Therefore, this layer contains semantic web services on different abstraction levels for control commands as well as for sensor data retrieval. To describe the services semantically, each service contains a semantic description such as proposed by OWL-S or the *Web Service Modeling Ontology (WSMO)* [32] respectively. In our concept, we use parts from both of them for semantic enrichment of the web services. The resulting service ontology is merged with the domain ontology to relate physical resources to their corresponding services. Finally, these web services are accessible using a web server that can be seen as an example of a *service layer* implementation. Thereby, each service has a *Service Grounding* that describes the access to the actual service of the *service layer*. In this regard, the *service layer* constitutes the connection between the *enactment layer* and *adaptation layer* with the *cyber-physical layer*, whereby the web server communicates with both upper layers and subsequently executes the desired web services. At the *enactment layer* and *adaptation layer*, AI methods for intelligent control can be developed that need access to the *Service Profile* for discovering services and to the *Service Model* for how the service works.

The invocation of a web service results in a concrete action in the *cyber-physical layer*, where the functionality of the web service is executed. As shown in Fig. 4, the controller that manages the corresponding physical manufacturing resource receives the request to execute a certain command, if necessary with special parameter settings. The controller actuates the managed device or measures the corresponding value of a sensor. Afterwards, the raw data or generally the response is transferred back to the web server, i.e., the sensor value or the message that the command has been executed both with a start and end time stamp.

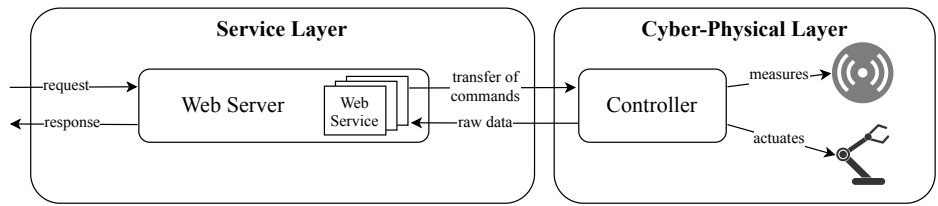

**Fig. 4.** Illustration of the Execution Sequence of a Web Service Invocation.

### 3.3 Semantic Web Services for a Physical Factory Model

Based on the presented architecture, a *top-down* approach has been applied to develop 67 basic SWSs that allow to use independently all functions of the physical devices of the complete shop floors presented in Sect. 2.2. The approach has been used for the identification of services and selection according to the level of abstraction corresponding to the use cases (see RQ 4). All in all, 336 SWSs with all parameter combinations have been derived in the *refinement* phase and modeled from these 67 SWSs (see RQ 7). During this phase, pre- and postconditions have been modeled as additional web services, results with included effects have been added, and all web services have been linked to the existing ontology (see RQ 2). For using AI planning techniques, the conditions and effects are additionally modeled with SPARQL expressions similar to Puttonen et al. [29]. The provided services are divided into services that perform an activity in the physical factory and into services that are offered for measuring sensor data. Furthermore, the services are hierarchically ordered into several more specific classes (see RQ 6). Table 1 illustrates an overview of the developed services arranged according to their corresponding physical resource and the possibility of parameterization. The number of different parameter settings for each service is given in brackets. By encapsulating services of different controllers of the cyber-physical layer, our service layer contributes to achieving interconnectivity and interoperability (see RQ 1). In addition to the controllers that run the simulation factory itself, Raspberry Pi´s offer services for retrieving RFID data. To describe the services semantically, we have re-modeled the important components of OWL-S and WSMO for our work and have especially tailored them to our simulation context. This means that we only have one service class, corresponding to the Service Profile in OWL-S, that describes the functionality of the service w.r.t.

**Table 1.** Overview of Semantic Web Services.

| Drilling Machine | High-Bay Warehouse | Milling Machine | Oven |
|---|---|---|---|
| calibrate (1) | calibrate (4) | calibrate (1) | calibrate (1) |
| drill (2) | changeBuckets (81) | mill (4) | burn (2) |
| transportFromTo (3) | store (10) | moveFromTo (6) | getMotorSpeed (1) |
| getMotorSpeed (3) | unload (10) | transportFromTo (6) | setMotorSpeed (1) |
| setMotorSpeed (3) | getMotorSpeed (4) | checkPosition (3) | resetAllMotors (1) |
| resetAllMotors (1) | setMotorSpeed (4) | getMotorSpeed (3) | statusOfLightBarrier (1) |
| capacitiveSensor (3) | resetAllMotors (1) | setMotorSpeed (3) | stateOfMachine (1) |
| statusOfLightBarrier (2) | statusOfLightBarrier (4) | resetAllMotors (1) | |
| stateOfMachine (1) | stateOfMachine (1) | statusOfLightBarrier (1) | |
| | getAmountOfStored Workpieces (1) | stateOfMachine (1) | |
| **Punching Machine** | **Sorting Machine** | **Vacuum Gripper Robot** | **Workstation Transport** |
| calibrate (1) | sort (20) | calibrate (4) | calibrate (1) |
| punch (2) | getMotorSpeed (1) | moveTo (9) | moveTo (2) |
| transportFromTo (3) | setMotorSpeed (1) | pickUpAndTransport (72) | pickUpAndTransport (2) |
| getMotorSpeed (3) | resetAllMotors (1) | checkPosition (9) | checkPosition (2) |
| setMotorSpeed (3) | statusOfLightBarrier (5) | getMotorSpeed (3) | getMotorSpeed (3) |
| resetAllMotors (1) | stateOfMachine (1) | setMotorSpeed (3) | setMotorSpeed (3) |
| capacitiveSensor (3) | | resetAllMotors (1) | resetAllMotors (1) |
| statusOfLightBarrier (2) | | stateOfMachine (1) | stateOfMachine (1) |
| stateOfMachine (1) | | | |

its parameters, inputs, outputs, preconditions, and results. Additionally, the description of postconditions is added from WSMO. These semantic descriptions can be used to determine what functionality the service provides, what requirements have to be fulfilled for execution, and how its successful execution can be verified (see RQ 5). Whereas most previous work updates the knowledge base and verifies the condition expressions based on the current real world state, we link to other semantic web services and use their response for our condition verification. This means that conditions are evaluated in near real-time based on sensor data that is accessed via a web service invocation (see RQ 3). The web service used to check the conditions can in turn require further web service invocations for condition verification (see RQ 9). To handle multiple parallel requests to one physical resource, a queue according to the First-In-First-Out principle is implemented and services are only executed by the web server when all required resources are not blocked by a previous request (see RQ 8). The goal of the queue is to store incoming web service requests to ensure that no conflicts between multiple requests occur during execution of different resources. After the service is carried out, we control the successful execution with postconditions that must be satisfied. If all postconditions are fulfilled, the successful response is given to the client.

In the following, we present one of the modeled semantic services in more detail. For this purpose, the *Vacuum Gripper Robot's (VGR) pick-up and transport service* is selected since it contains most of the annotated semantic elements such as pre- and postconditions as well as results. Listing 1 illustrates the corresponding OWL description in XML syntax and Fig. 5 the semantic annotations and their relationships as a graph.

```xml
1  <owl:NamedIndividual rdf:about="http://iot.uni-trier.de/FTOnto#
       Service_VGR_Pick_Up_And_Transport_VGR_1_Start_Sink_1_End_Oven">
2     <rdf:type rdf:resource="http://iot.uni-trier.de/FTOnto#
          ServiceVGRPickUpAndTransport"/>
3     <FTOnto:hasDescription rdf:resource="http://iot.uni-trier.de/FTOnto#
          Service_Description_VGR_Pick_Up_And_Transport"/>
4     <FTOnto:hasPrecondition rdf:resource="http://iot.uni-trier.de/FTOnto#
          Precondition_OV_1_State_Of_Machine_Ready"/>
5     <FTOnto:hasPrecondition rdf:resource="http://iot.uni-trier.de/FTOnto#
          Precondition_OV_1_Status_Of_Light_Barrier_5_Interrupted_False"/>
6     <FTOnto:hasPrecondition rdf:resource="http://iot.uni-trier.de/FTOnto#
          Precondition_SM_1_Status_Of_Light_Barrier_6_Interrupted_True"/>
7     <FTOnto:hasPrecondition rdf:resource="http://iot.uni-trier.de/FTOnto#
          Precondition_VGR_1_State_Of_Machine_Ready"/>
8     <FTOnto:hasPrecondition rdf:resource="http://iot.uni-trier.de/FTOnto#
          Precondition_WT_1_Check_Position_Oven_False"/>
9     <FTOnto:hasPostcondition rdf:resource="http://iot.uni-trier.de/FTOnto#
          Postcondition_OV_1_Status_Of_Light_Barrier_5_Interrupted_True"/>
10    <FTOnto:hasURL rdf:datatype="http://www.w3.org/2001/XMLSchema#string">
          http://127.0.0.1:5000/vgr/pick_up_and_transport?machine=vgr_1&
          start=sink_1&end=oven"/>
11  </owl:NamedIndividual>
```

**Listing 1.** XML Representation of Service Pick Up and Transport from Vacuum Gripper Robot.

In this context, green rectangles with rounded corners represent data properties, violet rectangles represent instances of classes that are in turn represented by orange ellipses. If invoked, this service fulfills the function of picking up a workpiece at sink one of the sorting machine, transporting it to the oven, and eventually dropping it off at the oven. Before the execution starts, the request is scheduled in the queue and as soon as no other request comes first, the execution of the service starts. At this point, access to the physical resource for other clients is blocked and intermediate requests are stored in the queue. The first part of the execution is the check of the preconditions. In this case, the SPARQL query as shown in Listing 2 is executed and returns five preconditions that must be fulfilled (see Fig. 5 or Listing 1).

```
1  PREFIX rdf: <http://www.w3.org/1999/02/22-rdf-syntax-ns#>
2  PREFIX owl: <http://www.w3.org/2002/07/owl#>
3  PREFIX rdfs: <http://www.w3.org/2000/01/rdf-schema#>
4  PREFIX xsd: <http://www.w3.org/2001/XMLSchema#>
5  PREFIX ftonto: <http://iot.uni-trier.de/FTOnto#>

7  SELECT ?service ?precondition ?preconditonCheckService ?checkURL ?
       requiredKeyInServiceResponse ?requiredValueInServiceResponse
8    {
9        ?service ftonto:hasURL "http://127.0.0.1:5000/vgr/pick_up_and_transport
             ?machine=vgr_1&start=sink_1&end=oven"^^xsd:string .
10       ?service ftonto:hasPrecondition ?precondition .
11       ?precondition ftonto:isCheckedBy ?preconditonCheckService.
12       ?preconditonCheckService ftonto:hasURL ?checkURL.
13       ?precondition ftonto:requiredKeyInServiceResponse ?
             requiredKeyInServiceResponse .
14       ?precondition ftonto:requiredValueInServiceResponse ?
             requiredValueInServiceResponse .
15   }
```

**Listing 2.** SPARQL Expression for Retrieving Preconditions.

For instance, the oven must be available and ready (see Listing 1 line 4) and the light barrier that monitors the target position of the transport must not be interrupted, because that indicates an empty storage space (see Listing 1 line 5). In particular, these preconditions refer to other semantic web services that perform the verification in near real-time. Therefore, complex reasoning within the knowledge base for precondition verification is not required. This could otherwise lead to considerable overhead for reasoning and possibly wrong, not real-time information for decision making. For providing real-time data, web services to retrieve the status of a sensor (e.g., a light barrier) are handled by a separate queue as the web services that initiate manufacturing operations. This division enables an immediate result even if the corresponding machine is still performing a manufacturing operation. The described principle is also applied to postconditions that are semantic web services too. The exemplary service contains one postcondition that checks whether the service has been executed successfully (see Listing 1 line 9). The postcondition checks whether the light barrier, which was not interrupted for the corresponding precondition, has now been interrupted, i.e., it is verified that the workpiece has been transported from the first sink of the sorting machine to the oven and thus the execution was successful. For each service, regardless of whether it is used as a pre- or postcondition or neither, the respective URL to invoke the service is represented as instance (see Listing 1 line 10 and Fig. 5).

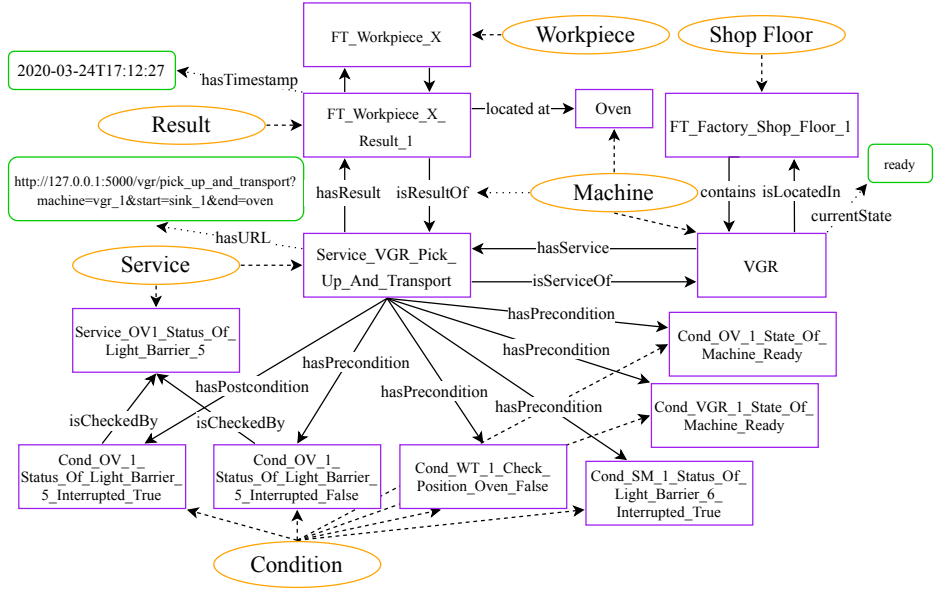

**Fig. 5.** Semantic Annotations of the Pick Up and Transport Service from Vacuum Gripper Robot as a Graph.

## 4  Evaluation

In this section, we evaluate our developed SWSs as part of the *Evaluation* phase of the ontology development methodology for usefulness for the described use cases in Sect. 2.3. Therefore, we have prototypically implemented the third use case by using the workflow management system Camunda[3]. Camunda is able to invoke web services by applying *Business Process Model and Notation (BPMN)* conform *Service Tasks*. We implement a small cyber-physical production workflow that transports a workpiece from the first sink of the sorting machine to the oven, burns it, and after a quality inspection by an employee, it is transported and stored in the high-bay warehouse. Figure 6 illustrates the described manufacturing process as BPMN workflow. It is possible to execute the workflow in the Camunda workflow engine. The corresponding web server receives the web service invocations and forwards the execution command to the controller of the specified physical device. Before execution, an implemented Python class checks if the preconditions and after execution the corresponding postconditions of the service are satisfied by using OWLReady[4] [15].

---

[3] https://camunda.com/

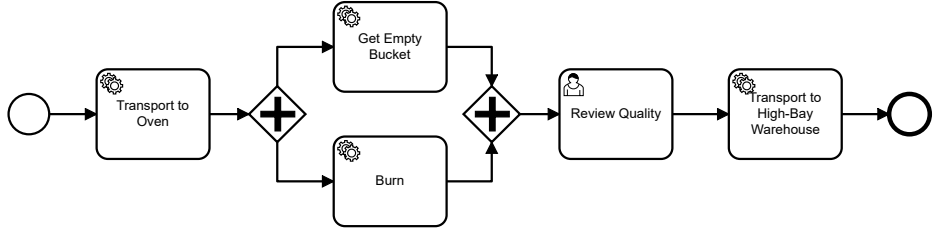

**Fig. 6.** A BPMN Workflow of a Manufacturing Process.

Furthermore, we check the developed SWSs for conformity and correctness by using the *OntOlogy Pitfall Scanner! (OOPS!)*[5]. In this process, no errors or inconsistencies have been detected.

## 5  Conclusion and Future Work

In this work, we present semantic web services for AI-based research in Industry 4.0. First, we describe use cases in which the use of semantic web services is valuable. Based on the described use cases, we derive requirements that should be met by semantic services. As a result of our research, we modeled 336 semantic web services based on standards such as OWL-S and WSMO. The semantic services are enriched with inputs, outputs, preconditions, results, and postconditions. In our evaluation, we exemplary show that the developed semantic services are suitable for using in a workflow management system to build valid cyber-physical production workflows. The developed services provide the foundation to support low-code applications [33].

In future work, we investigate the described use cases further to enhance workflow flexibility in cyber-physical systems. We think that *Process-Oriented Case-Based Reasoning (POCBR)*, for example, could be used for this purpose because it has already shown great potential in various other domains. Additionally, automated planning techniques can be used to further increase workflow flexibility in Industry 4.0. With the future application, the semantic web services are continuously improved according to their environment and requirements (see *Application & Evolution* phase). The developed semantic web services are available for download at http://iot.uni-trier.de.

**Acknowledgments.** This work is funded by the German Research Foundation (DFG) under grant No. BE 1373/3-3. The basic components of the semantic web services were developed and implemented in a student research project at Trier

---

[4] https://pypi.org/project/Owlready2/
[5] http://oops.linkeddata.es/

University by Felix Reither and Julian Sawatzki. Revisions of the services have been carried out by the student research assistant Marcel Mischo.